\pdfoutput=1

\documentclass[11pt]{article}

\usepackage[]{ACL2023}

\usepackage{booktabs}
\usepackage{graphicx}
\usepackage{tabularx}
\usepackage{makecell}
\usepackage{multicol}

\usepackage{times}
\usepackage{latexsym}

\usepackage[T1]{fontenc}

\usepackage[utf8]{inputenc}

\usepackage{microtype}

\usepackage{inconsolata}

\title{Unknown Script: Impact of  
 Script on Cross-Lingual Transfer  }

\author{Wondimagegnhue Tsegaye Tufa, Ilia Markov, Piek Vossen \\Vrije Universiteit Amsterdam \\ 
De Boelelaan 1105, 1081 HV Amsterdam, The Netherlands \\
\{w.t.tufa, i.markov, p.t.j.m.vossen\}@vu.nl\\}

\begin{document}
\maketitle

\begin{abstract}
Cross-lingual transfer has become an effective way of transferring knowledge between languages. In this paper, we explore an often-overlooked aspect in this domain: the influence of the source language of a language model on language transfer performance. We consider a case where the target language and its script are not part of the pre-trained model. We conduct a series of experiments on monolingual and multilingual models that are pre-trained on different tokenization methods to determine factors that affect cross-lingual transfer to a new language with a unique script. Our findings reveal the importance of the tokenizer as a stronger factor than the shared script, language similarity, and model size.
\end{abstract}


\section{Introduction}
\begin{figure}[ht]
  \centering
  \resizebox{0.5\textwidth}{!}{%
    \includegraphics{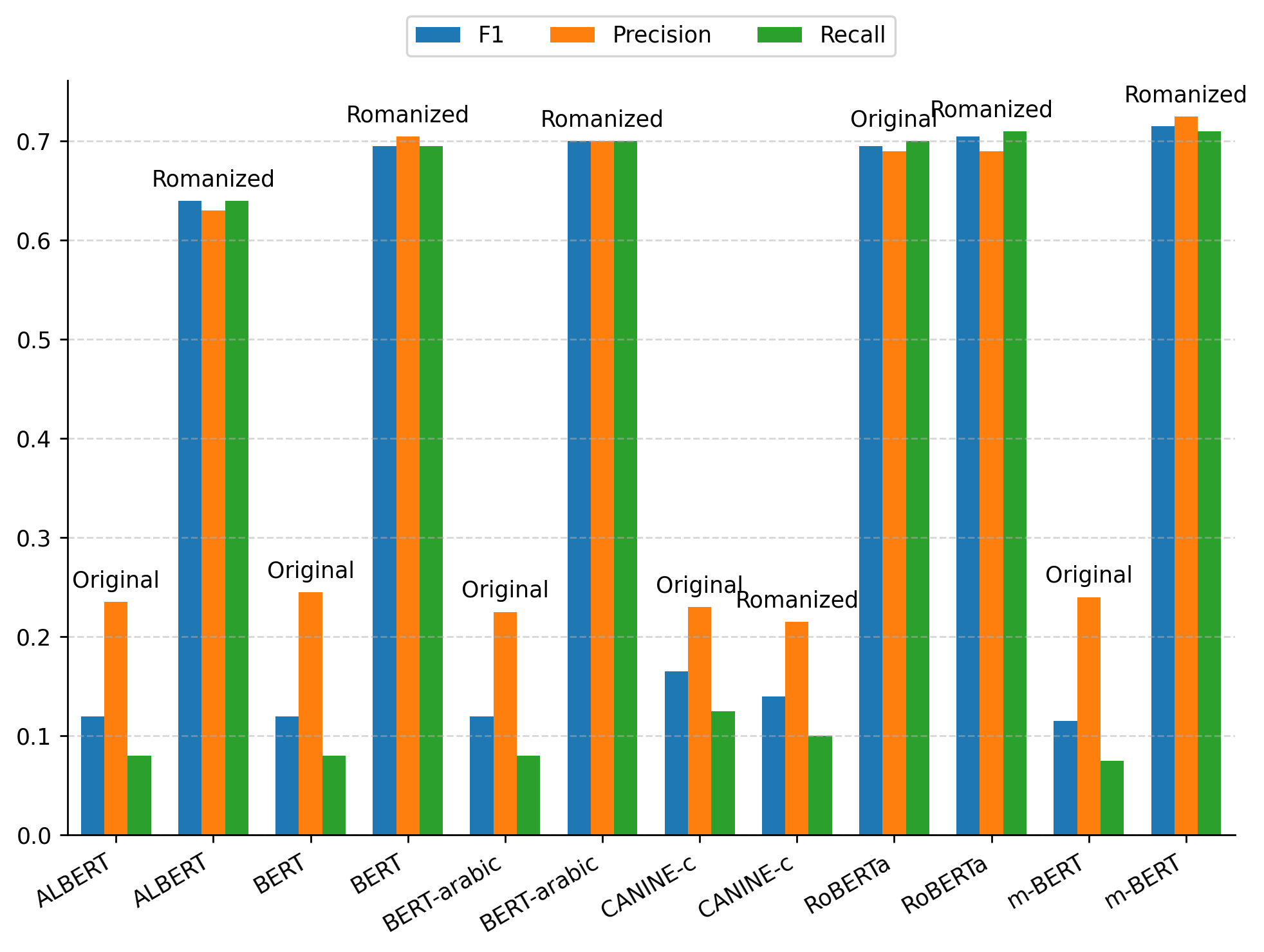}
  }
  \caption{We analyze the effect of script and tokenizer on cross-lingual transfer on a target language with a new script. We select six monolingual and multilingual models pre-trained using sub-word tokenizers and character tokenizers. We fine-tune these models on the NER and POS tasks in the original script (FIDEL) and the romanized version (Latin). We observe that RoBERTa has better cross-lingual transfer in both the original script and the romanized version. We also find that romanization is strongly beneficial in all cases of subword-based models (ALBERT, BERT,m-BERT). Additionally, fine-tuning Arabic-BERT, which is typologically similar to our target language (Amharic), provides no advantage. We employ the base version of the models across all cases to ensure a fair comparison. The reported F1-score is averaged over five runs, with a standard deviation ranging between -0.003 and 0.009. }
  \label{fig:main}
\end{figure}
The dominant Natural Language Processing (NLP) approach nowadays involves cross-lingual transfer using pre-trained monolingual and multilingual language models. In line with this trend, numerous monolingual models have been released for various languages \cite{devlin-etal-2019-bert,cañete2023spanish,antoun-etal-2020-arabert}. Multilingual models, which are trained on 100 or more languages, have also been introduced, such as XLM-R~\cite{conneau2019unsupervised} and m-BERT~\cite{devlin-etal-2019-bert}. 

Despite the advancements in the development of language models for high-resourced languages, the vast majority of the world's languages remain excluded from these models. Out of over 6,500 spoken languages globally, less than 2\% are represented in the current models, leaving the rest unseen and unaccounted for in the current language processing technology \cite{muller-etal-2021-unseen,hammarstrom2016linguistic,joshi-etal-2020-state}. Training a model for each of these languages is impractical due to substantial data and computational resource requirements. Several alternative approaches have been proposed, such as zero-shot transfer \cite{pires-etal-2019-multilingual,conneau2019unsupervised}, language adapters \cite{pfeiffer-etal-2020-mad}, and extending multilingual models \cite{conneau2019unsupervised,devlin-etal-2019-bert}. 

While such methods require fewer resources in the target language, they often yield sub-optimal results \cite{lauscher-etal-2020-zero,pfeiffer-etal-2021-unks}. For example, using even a small amount of labeled data in the target language has been shown to be more effective than a zero-shot transfer ~\cite{lauscher-etal-2020-zero}.
\\

The effectiveness of cross-lingual transfer is influenced by several factors, such as language similarity between the source and target languages, the size of the pre-trained model, and the quality and amount of the pre-training and fine-tuning data \cite{muller-etal-2021-unseen,pires-etal-2019-multilingual,DBLP:journals/corr/abs-2002-03518,wu-dredze-2019-beto}. 

In this paper, we explore which factors determine cross-lingual transfer performance for a new language with a unique script. We consider a challenging case where the target language is not part of the pre-trained model, and the script has also not been seen in pre-training. Our analysis targets three main factors: language similarity, tokenization methods, and script. We design an experiment by varying these factors. We evaluate a range of existing monolingual and multilingual models, specifically choosing those trained on typologically related or typologically distant languages. Furthermore, we select models trained using various tokenizers, allowing us to assess how these choices impact cross-lingual performance for a language with a unique script. 
We focus on two main research questions:
\begin{enumerate}
  \item To what extent does the script of a source language influence cross-lingual transfer to a new language in monolingual and multilingual models? 
   \item To what extent does the tokenizer influence cross-lingual transfer to a new language in monolingual and multilingual models?
\end{enumerate}
Figure \ref{fig:main} shows the results of our experiment. Our analysis shows that RoBERTa has better cross-lingual transfer irrespective of the script.
However, romanization strongly affects cross-lingual transfer for models pre-trained using sub-word tokenizers in monolingual and multilingual settings.   
We make our code available.\footnote{\url{https://github.com/cltl/unkown_script/tree/main}}

\section{Related Work}
Under-resourced languages display considerable variation in several aspects \cite{joshi-etal-2020-state}. First, the amount of data for these languages varies greatly. Secondly, many of these languages use scripts different from the Latin script \cite{muller-etal-2021-unseen,joshi-etal-2020-state}. Finally, regarding linguistic characteristics, these languages often have distinct morphological and syntactic properties, especially when compared to high-resourced Indo-European languages. 

Recently, cross-lingual transfer has become an effective method for extending the capabilities of pre-trained monolingual and multilingual models for various languages. In this section, we present an overview of studies exploring cross-lingual transfer. 

\paragraph{Multilingual models and language adapters}
Multilingual models enable transfer between high-resource and low-resource languages \cite{conneau2019unsupervised,devlin-etal-2019-bert}. However, they suffer from the `curse of multilingualism' and interference between languages \cite{conneau2019unsupervised,wang2020negative}, where the model's effectiveness decreases as the number of languages increases due to the parameter limit of the model. Language adapters address these challenges by storing language-specific knowledge of each language in dedicated parameters \cite{pfeiffer-etal-2020-mad}. This increases the capacity of a multilingual model without introducing interference between languages. These methods are, however, not directly applicable to languages that use scripts not covered in the training data of these models \cite{pfeiffer-etal-2021-unks}.

\paragraph{Zero-shot and few-shot transfer} In zero-shot transfer, a fine-tuned model on a resource-rich source language is directly applied to a resource-poor target language~\cite{pires-etal-2019-multilingual,conneau2019unsupervised}. While this method is appealing, it often yields sub-optimal results \cite{lauscher-etal-2020-zero,pfeiffer-etal-2021-unks}. Alternatively, using even a small amount of labeled data in the target language (few-shot transfer) has shown to be more effective~\cite{lauscher-etal-2020-zero}. It remains unclear what factors determine this effect and to what extent.

\paragraph{Factor analysis}
The success of cross-lingual transfer is impacted by various factors \cite{muller-etal-2021-unseen,pires-etal-2019-multilingual,DBLP:journals/corr/abs-2002-03518,wu-dredze-2019-beto}. \citet{muller-etal-2021-unseen} demonstrated that the performance of transfer can significantly differ based on factors such as the script of the language, the amount of available data, and the relationship between source and the target language. However, the literature concerning the effect of script and tokenizer is mixed. For example, \citet{muller-etal-2021-unseen} identifies script as the most crucial factor affecting transfer performance and shows that transliterating a script to the Latin script enhances the effectiveness of cross-lingual transfer. Contrary to this,  \citet{artetxe2019cross} and \citet{k2020crosslingual} show that script and lexical overlap are less relevant and that large monolingual models learn semantic abstractions that are generalizable to other languages. A similar mixed result has been reported when examining the effect of tokenizers in cross-lingual transfer. \citet{rust-etal-2021-good} show that tokenizers are a crucial factor in the success of cross-lingual transfer for multilingual models, while \citet{wu2022oolong} report it as less important. The analysis of \cite{muller-etal-2021-unseen} covers multilingual models, while \citet{wu2022oolong} focuses only on an English model.

While similar to the approach of \citet{muller-etal-2021-unseen}, our study takes a different direction. Instead of analyzing the performance of multiple languages with a single multilingual model, we focus on one language that is unique in its script and not covered by existing models. We select Amharic as our target language. Amharic is a Semitic language and is classified as morphologically complex. It has a unique script, distinct from Latin and Arabic alphabets, with no shared characters. Additionally, it is categorized as a Class 2 under-resourced language, indicating a significant lack of data and tools for language processing \cite{joshi-etal-2020-state,DBLP:journals/corr/abs-2103-11811}.
We evaluate a range of existing monolingual and multilingual models, specifically choosing those trained on languages either closely related to or distinct from Amharic. Furthermore, we select models trained using various tokenizers, allowing us to assess how these choices impact model performance for a language with a unique script. In this way, we measure the impact of different factors on cross-lingual transfer.


\section{Methodology}
We experiment with a few-shot setting in which we fine-tune pre-trained models on downstream tasks. The target language and script are not part of the pre-trained models we explore. Our few-shot setup follows a standard setting: we take an existing base model, fine-tune it, and test it on the original target language. 

\paragraph{Language}
We experiment with Amharic as our target language. According to the language classification system presented in \cite{joshi-etal-2020-state}, Amharic is categorized as a Class 2 under-resourced language. It possesses a unique script and is characterized by its morphological complexity. 
We select two source languages for our analysis: English, which is typologically distant from Amharic, and Arabic, which is typologically related. Both English and Arabic of these languages do not share a script with Amharic. We use the original Fidel script and the romanized version for our fine-tuning and evaluation.

\subsection{Task and model}
Table \ref{tab:model_used} shows a summary of the pre-trained models we use and their corresponding tokenizers. The selection includes monolingual models trained in English and Arabic and various multilingual models. 

\begin{table*}[ht]
\small
\centering
\begin{tabular}{@{}lcccc@{}}
\toprule
Model & Tokenizer & Model-Type & Language & Model Size \\ 
\midrule
BERT \cite{devlin-etal-2019-bert} & WordPiece & Monolingual & English & 110M  \\  
RoBERTa \cite{liu2021robustly} & BPE & Monolingual & English & 125M  \\  
ALBERT \cite{devlin-etal-2019-bert} & SentencePiece & Monolingual & English & 12M \\  
BERT-base-arabic \cite{antoun-etal-2020-arabert} & WordPiece & Monolingual & Arabic & 110M \\  
CANINE-c \cite{Clark_2022} & Character & Multilingual & Multiple & 121M  \\   
m-BERT \cite{devlin-etal-2019-bert} & WordPiece & Multilingual & Multiple & 110M  \\ 
\bottomrule
\end{tabular}
\caption{Overview of the models we use, tokenizers, model types, languages, and the model size.}
\label{tab:model_used}
\end{table*}

\begin{table*}[h]
\small
\centering
\begin{tabular}{@{}lccccccccccccc@{}}
\toprule
Model & \multicolumn{3}{c}{Fidel-NER} & \multicolumn{3}{c}{Latin-NER} & \multicolumn{3}{c}{Fidel-POS} & \multicolumn{3}{c}{Latin-POS} \\ 
\cmidrule(r){2-4} \cmidrule(l){5-7} \cmidrule(l){8-10} \cmidrule(l){11-13}
& F1 & P & R & F1 & P & R & F1 & P & R & F1 & P & R \\ 
\midrule
RoBERTa-base & \textbf{0.57} & 0.59 & 0.55 & 0.55 & 0.53 & 0.56 & \textbf{0.82} & 0.79 & 0.85 & \textbf{0.86} & 0.85 & 0.86 \\
CANINE-c & 0.12 & 0.13 & 0.1 & 0.09 & 0.12 & 0.07 & 0.21 & 0.33 & 0.15 & 0.19 & 0.31 & 0.13 \\
BERT-base & - & - & - & 0.57 & 0.6 & 0.55 & 0.24 & 0.49 & 0.16 & 0.82 & 0.81 & 0.84 \\
m-BERT  & - & - & - & \textbf{0.59} & 0.62 & 0.57 & 0.23 & 0.48 & 0.15 & 0.84 & 0.83 & 0.85 \\
BERT-arabic & - & - & - & \textbf{0.59} & 0.6 & 0.58 & 0.24 & 0.45 & 0.16 & 0.81 & 0.8 & 0.82 \\
ALBERT-base & - & - & - & 0.5 & 0.5 & 0.49 & 0.24 & 0.47 & 0.16 & 0.76 & 0.76 & 0.79 \\
\bottomrule
\end{tabular}
\caption{Results of few-shot experiments where we fine-tune different models on NER and POS tasks with Fidel and romanized script. The empty cells show that we do not observe a decrease in the loss. We fine-tune all our models for 25 epochs. The F1 score is averaged over five runs with a standard deviation between 0.003 and 0.009. The highest F1 score for each script is highlighted in bold.}
\label{tab:result}
\end{table*}

\paragraph{Task and Datasets}
We experiment with two tasks: Named Entity Recognition (NER) and part-of-speech (POS) tagging. For the NER task, we use MasakhaNER \cite{DBLP:journals/corr/abs-2103-11811}, and for POS, we use the Amharic-POS dataset from \cite{DBLP:journals/corr/abs-2106-07241}. The MasakhaNER dataset is a publicly accessible resource for the NER task in ten African languages. This dataset has four types of entity labels: PER (Person), ORG (Organization), LOC (Location), and DATE (Date). It has 1,750 training, 250  validation, and 500 test instances. The POS dataset contains 218K sentences with 18 POS tags. We sample 2.5K examples, using 1,750 sentences for training, 250 for validation, and 500 for testing.

\paragraph{Tokenizers}
Our model selection encompasses the most widely used tokenizers: SentencePiece \cite{kudo-richardson-2018-sentencepiece}, BPE \cite{sennrich-etal-2016-neural}, character-based tokenizer \cite{Clark_2022} and WordPiece \cite{schuster2012japanese}.

\paragraph{Fine-tuning}
We fine-tune all models on two tasks. Our fine-tuning is challenging because our target language is not seen during pre-training and has a unique non-Latin script. The experiment is designed to test two capabilities. First, we evaluate whether the models we are testing enable cross-lingual transfer to a new language and script unseen in pre-training. This experiment is intended to shed light on cross-lingual transfer to a target language that does not share a script with the source language. Second, we explore how different tokenization methods might facilitate cross-lingual transfer.

\section{Results and Analysis}
Table \ref{tab:result} shows the result for the few-shot setting on NER and POS fine-tuned on the original Amharic script (Fidel) and its romanized version. 
The RoBERTa-base model stands out, showing robust performance across both scripts, with a marginal preference for the Fidel script for the NER task and the Latin script for the POS task. Models such as BERT-base, m-BERT, BERT-Arabic, and ALBERT-base fail to recognize entities entirely in the Fidel script but show some capabilities with the Latin script. This pattern persists across both tasks, although the POS task has less variation.

\paragraph{Effect of the script }
The difference in the obtained results for Fidel and romanized versions highlights the script's effect on model performance. Models pre-trained on data predominantly in the Latin script struggle significantly with tasks in the Fidel script, as shown by the drastically lower F1 scores for most models trained on the Fidel script compared to the Latin script. This suggests a strong bias towards the script used during the training phase, with models performing better on scripts they have seen before. 
This is in line with the result reported by \citet{purkayastha2023romanizationbased,muller-etal-2021-unseen}, which shows that the romanization of unknown script boosts transfer performance in multilingual models. However, we also observe this effect in all of the monolingual models we test. The RoBERTa model seems to be an exception, showing a good performance before romanization, though romanization also slightly improves its performance. 

\paragraph{Language relatedness}
English BERT-base and Arabic BERT can be compared directly since they are trained with similar training objectives, model sizes, and tokenizers. We observe a mixed result, with BERT-Arabic performing slightly better on the NER task but showing a lower score on the POS task.
\paragraph{Model size and tokenizers}
A plausible explanation for the performance variation between RoBERTa and the other models could be attributed to the model size and tokenizer. RoBERTa-base is the largest model with 125 million parameters. However, the performance does not consistently correlate with the model size. The other possible explanation is the tokenizer used. RoBERTa is trained using BPE over raw bytes instead of Unicode characters. The results show that the BPE representation enables the model to leverage knowledge that benefits the downstream tasks, even if the script is not included in the model's vocabulary.

\section{Conclusion}
In this paper, we explore cross-lingual transfer in less explored but challenging settings where the target language is not seen during pre-training and has a unique non-Latin script. 
Our analysis shows considerable differences in cross-lingual transfer performance among various models, possibly attributable to two key factors: the size of the pre-trained model and the specific tokenizer used during pre-training. The model's size could impact its ability to capture and generalize across multiple languages, including a language distant from the pre-trained model's language. In light of recent studies that show the importance of tokenizers in cross-lingual transfer for under-resourced languages, we show that a choice of tokenizer plays a role in facilitating cross-lingual transfer. 

\section{Limitations}
In our analysis, we intend to control for various factors that could influence the comparative performance of different models. However, residual differences in model parameters and the extent of pre-training data may have contributed to the observed disparities in the obtained results. Furthermore, our study did not involve training a model from scratch with fixed architecture, parameters, data, and domain while varying only the tokenizer or the model size. This limitation precludes a definitive conclusion about the isolated effect of the tokenizer on model performance. Hence, we identify the need for additional research where these elements are carefully controlled. Such an experiment would enable a more robust understanding of the tokenizer's role and interaction with other model characteristics in cross-lingual transfer learning.

\bibliography{custom}
\bibliographystyle{acl_natbib}

\end{document}